\documentclass[10pt,twocolumn,letterpaper]{article}

\usepackage[final]{cvpr}
\usepackage[colorlinks=true, linkcolor=black, urlcolor=black, citecolor=black]{hyperref}
\usepackage{times}
\usepackage{epsfig}
\usepackage{graphicx}
\usepackage{amsmath}
\usepackage{amssymb}

\usepackage{booktabs}
\usepackage{threeparttable}
\usepackage{multirow} 
\usepackage{makecell}
\usepackage{braket}
\usepackage{pifont}
\newcommand{\hhline}{\noalign{\vskip 1pt}\hline\noalign{\vskip 1pt}}

\def\ijcbPaperID{4} 

\begin{document}
\title{Latent Fingerprint Matching via Dense Minutia Descriptor}

\author{Zhiyu Pan \and Yongjie Duan \and Xiongjun Guan \and Jianjiang Feng\thanks{Jianjiang Feng is the corresponding author. \\ \indent This work was supported in part by the National Natural Science Foundation of China under Grant 62376132.} \and Jie Zhou \and Department of Automation, Tsinghua University, China\\
\tt\small \{pzy20, dyj17, gxj21\}@mails.tsinghua.edu.cn \\ \tt\small \{jfeng, jzhou\}@tsinghua.edu.cn}



\maketitle
\thispagestyle{empty}

\begin{abstract}
   Latent fingerprint matching is a daunting task, primarily due to the poor quality of latent fingerprints. In this study, we propose a deep-learning based dense minutia descriptor (DMD) for latent fingerprint matching. A DMD is obtained by extracting the fingerprint patch aligned by its central minutia, capturing detailed minutia information and texture information. Our dense descriptor takes the form of a three-dimensional representation, with two dimensions associated with the original image plane and the other dimension representing the abstract features. Additionally, the extraction process outputs the fingerprint segmentation map, ensuring that the descriptor is only valid in the foreground region. The matching between two descriptors occurs in their overlapping regions, with a score normalization strategy to reduce the impact brought by the differences outside the valid area. Our descriptor achieves state-of-the-art performance on several latent fingerprint datasets. Overall, our DMD is more representative and interpretable compared to previous methods. Corresponding code is available at \href{https://github.com/Yu-Yy/DMD}{https://github.com/Yu-Yy/DMD}.
\end{abstract}

\begin{figure}[!h]
	\centering
	\subfloat[\label{fig:1D_compare}]{\includegraphics[width=0.404\linewidth]{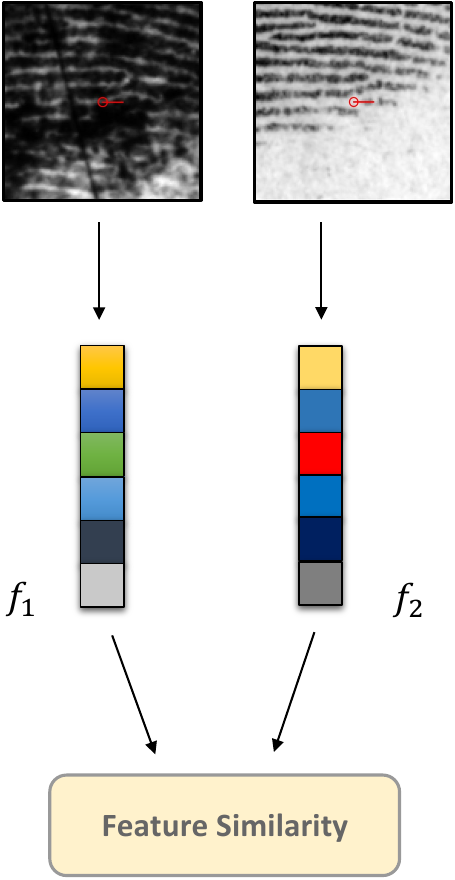}} 
	\subfloat[\label{fig:MDD_compare}]{\includegraphics[width=0.44\linewidth]{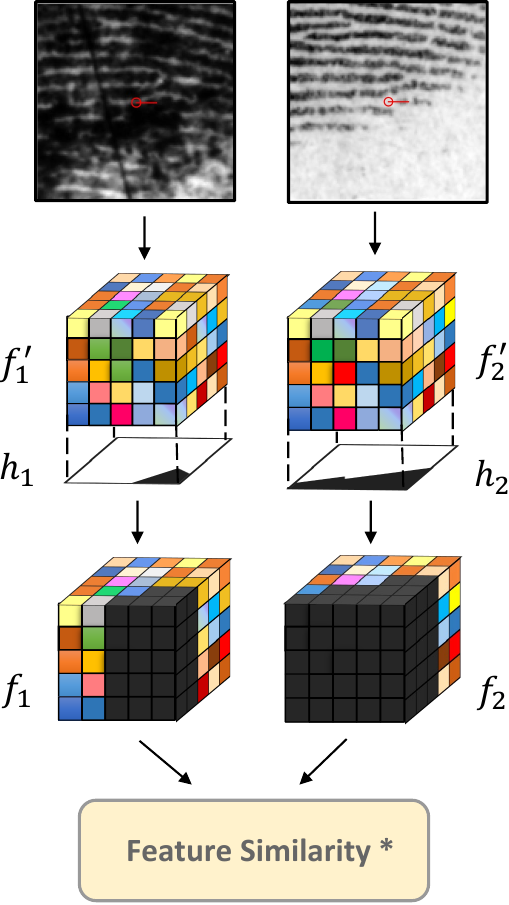}}
	\caption{Compared with (a) one-dimensional minutia descriptor, (b) our Dense Minutia Descriptor (DMD) is a three-dimensional representation, and explicitly considers the overlapping area for score normalization. Score normalization is denoted as *. }
	\label{fig:MDD_comparison}
\end{figure}

\section{Introduction}
   Fingerprints found at crime scenes, often referred to as latent fingerprints, are crucial for identifying suspects. Considering the global reliance of law enforcement agencies on latent fingerprint recognition technology \cite{maltoni2022handbook}, the inherent poor quality of such fingerprints—with indistinct ridge lines—necessitates professional examiner annotations in forensic investigations. Yet, these annotations are susceptible to discrepancies due to variations among examiners \cite{Cao2017AutoLatent}. Thus, the development of an automated latent fingerprint recognition and matching system would significantly bolster the ability of law enforcement agencies to solve crimes. 

   Due to the complex nature of latent fingerprint acquisition, ridge lines are often blurred and may be subject to background noise interference. As a result, many researchers have focused on effectively extracting or enhancing the level-1 (orientation field, frequency map, etc. \cite{DictMethod,LocalDict,Duan2021OrientField}) and level-2 (ridge skeleton map, minutiae, etc. \cite{Joshi2019GanEnhance, Huang2020GanEnhance,Zhu2023FingerGAN, Lu2016MinutiaDeepLearn,Darlow2017MinutiaDeepLearn,Tang2017LatentMinutia, Liu2021LatentSegEnhance}) features of latent fingerprints. 
   These methods have significantly enhanced the matching performance of conventional fingerprint matching techniques; however, these feature extraction steps may introduce noisy features or destroy original features, underscoring the need for latent fingerprint matching algorithms to exhibit robustness against the challenges posed by low-quality fingerprints.

   Given the limitations of handcrafted features \cite{FingerCode, Feng2008combining, Cappelli2011FastAccurate, Sankaran2011LatentMatch} in adapting to diverse fingerprint types and low quality situations, deep learning has been explored for abstract feature extraction in fingerprint matching. These methods are categorized into fixed-length descriptors and minutia-based representations. The former encodes fingerprints into fixed-length vectors, enhancing indexing efficiency, with approaches like multi-scale descriptors \cite{song2017fingerprint,gu2022latent} and integrating minutiae with texture features \cite{song2019aggregating, DeepPrint, Wu2022MinutiaeAwarely}. Recent works have also utilized Vision Transformer's potent extraction abilities for more comprehensive descriptors \cite{grosz2022minutiae, Grosz2023AFRNet}. However, these methods struggle with the nuanced description of latent fingerprints, which challenge single-vector representations due to their interference propensity. Additionally, fingerprint pose alignment \cite{jaderberg2015spatial, Duan2023FingerPose} necessary for most fixed-length techniques is compromised by the blurriness and incompleteness of latent fingerprints.

   Minutia-based fingerprint matching techniques hinge on aligning fingerprint images using each minutia's location and direction to subsequently extract local patch features. This approach, inherently resistant to overall fingerprint pose changes, provides superior accuracy to most fixed-length methods, albeit less efficiently \cite{grosz2023latent}. Works by \cite{cappelli2010minutia, medina2016latent, MinNet} have concentrated on encoding minutia relationships within a patch, designating one minutia as the anchor point, with {\"O}zt{\"u}rk et al. \cite{MinNet} incorporating CNNs for concurrent texture feature encoding. Beyond using minutiae as anchors, Cao et al. \cite{Cao2017AutoLatent, Cao2020E2E} integrated orientation fields as dense anchors for comprehensive depictions, termed virtual minutiae or texture templates \cite{Cao2018TextureTemplate}. Similarly, Gu et al. \cite{Gu2021LatentRegis} applied dense uniform sampling points as anchors to learn relative patch alignment and descriptor extraction.

   In this study, we introduce a deep-learning representation termed Dense Minutia Descriptor (DMD) which is representative and interpretable.
   Our approach diverges from the conventional use of one-dimensional deep representations, instead employing a dense descriptor in a three-dimensional form.
   This format not only retains spatial relations intrinsic to the original image, enhancing interpretability, but also aligns closely with the actual image structure, where the two dimensions represent a coarse image mapping and the third encodes texture features in depth.
   Our method's interpretability facilitates direct comparison between descriptors, mirroring specific local correspondences of the source images. To further refine matching precision, our network generates a segmentation map that isolates overlapping regions, thereby reducing background noise in descriptor comparisons. The matching between DMDs is considered only in their overlapping region.
   Drawing inspiration from \cite{Iris}, we also incorporate a matching score normalization technique based on the overlapped area, minimizing the influence of the area of overlapping region. The comparison between DMDs matching and other methods is shown in Figure \ref{fig:MDD_comparison}. Architecturally, our model adopts a dual-branch system akin to that of Engelsma et al. \cite{DeepPrint}, isolating the extraction of texture and minutiae-specific features to bolster fingerprint recognition accuracy.

   In this study, we conduct the experiments on two most commonly used latent datasets, NIST SD27 \cite{garris2000nist} and NIST SD302 Latent subset (N2N Latent) \cite{fiumara2019nist}. To thoroughly validate the effectiveness of the descriptors, we do not employ any preprocessing of the fingerprint images like image enhancement. The experiments demonstrate that our method outperforms other deep-learning based descriptor methods \cite{Cao2020E2E,MinNet}, conventional well-designed descriptor \cite{cappelli2010minutia}, and Commercial Off-The-Shelf (COTS) method \cite{nist2020verfinger}. Besides, DMD maintains good performance even after binarization, thus indicating its potential for practical applications as an automated fingerprint recognition system.

\begin{figure*}[!h]
	\centering
	\includegraphics[width=0.9\linewidth]{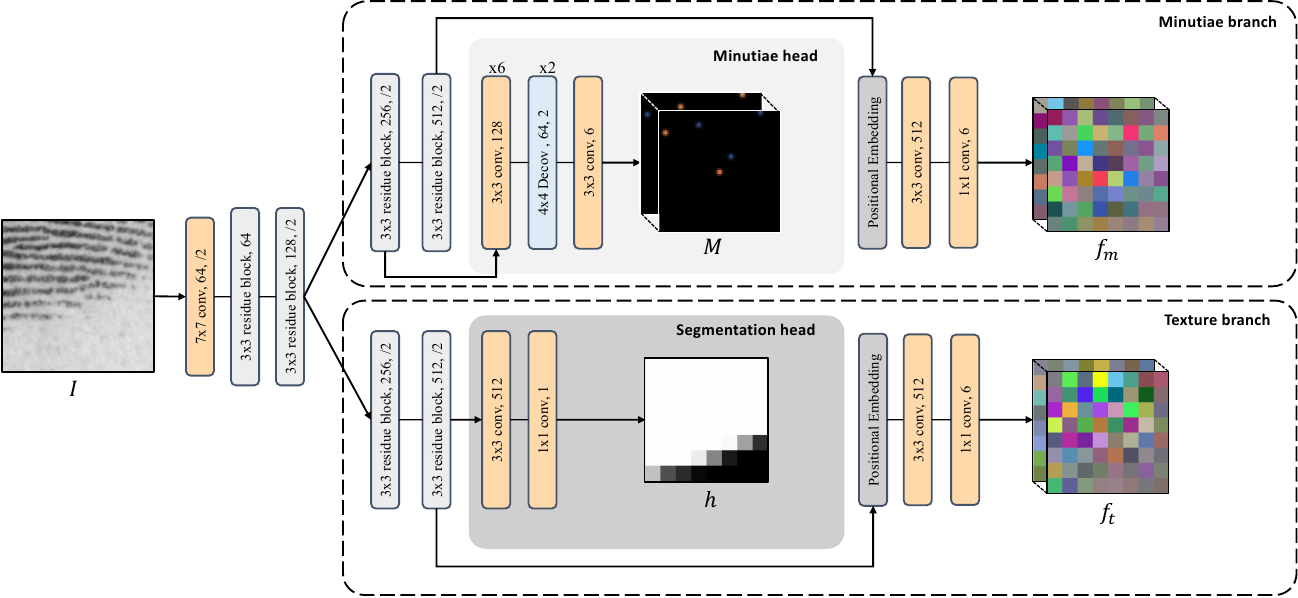}
	\caption{The detailed structure of our DMD extraction network. The content boxes display operation names, output channels, and spatial scales separated by commas. The third one is omitted if scale equals 1.}
	\label{fig:network}
\end{figure*}

\section{Method}
\subsection{Descriptor Extraction Network}
The basic backbone architecture is modified from ResNet-34 \cite{he2016deep} by removing the first max pooling layer for preserving the details of fingerprint ridges. Furthermore, we design a dual-branch structure which is split at the second residue block sets. One branch is tailored to produce a texture descriptor, complemented by a segmentation map as an auxiliary output. Correspondingly, the second branch is focused on generating a minutiae descriptor, alongside a minutiae map, also serving as an auxiliary output. In order to augment the network's ability to incorporate spatial information during comparing descriptors, we employ a 2D positional embedding \cite{vaswani2017attention} with the well-known sinusoidal form at each branch. The final Dense Minutia Descriptor (DMD) is the concatenation of two descriptors timed by the segmentation map. The overall structure is shown in Figure \ref{fig:network}.

\textbf{Texture Descriptor}.
At the texture branch, it outputs the segmentation map $h$ with auxiliary 2D convolution layers denoted as segmentation head. It enables a heightened focus on the distribution of the ridge lines. Additionally, the segmentation map $h$ is essential for DMD and serves a critical role in matching score normalization. The texture descriptor head shares the same feature map from the last residue block sets as the segmentation head. Consequently, we obtain the texture descriptor $f_\text{t}$ and the segmentation map $h$, where  $h\in\mathbb{R}^{1 \times 8 \times 8} $ and $f_\text{t}\in\mathbb{R}^{C \times 8 \times 8}$, with $C$ indicating the depth dimension.

\textbf{Minutiae map}.
Considering the large number and complex configure of minutiae within fingerprint images, we conceptualize the distribution of minutiae's position and orientation to a 6-channel 3D heatmap called minutiae map inspired by \cite{FingerNet,DeepPrint}. Here, two dimensions correspond to the image plane, while the third dimension encompasses angles ranging from $0^\circ$ to $360^\circ$. Each minutia is depicted using a Gaussian distribution, characterized by a variance of $\sigma^2$, centered around its specific position and orientation denoted by $(x,y,\theta)$. For our specific application, we have chosen to set the parameter $\sigma$ equal to 1.

\textbf{Minutiae Descriptor}.
Minutia is the detailed feature (level-2) of fingerprint, and hence we extract the feature of penultimate residue block sets to feed the minutiae. The minutiae head is composed by sets of 2D convolution layers and 2D Deconvolution layers to resume the high resolution minutiae map. It enables this branch focus more on the minutiae-related feature. The minutiae descriptors head is directly connected to the last result block sets. Consequently, we can obtain the minutiae descriptor $f_\text{m}\in\mathbb{R}^{C\times8\times8}$ and minutiae map $M\in\mathbb{R}^{6\times64\times64}$, with the $C$ is the same as the one in texture descriptor.

Finally, we can get the dense descriptor $f\in\mathbb{R}^{2C\times8\times8}$ by concatenating two types descriptors and multiplexing the segmentation map $h$ as
\begin{equation}
   \label{eq:generate_desc}
   f = (f_\text{t} \oplus f_\text{m}) \odot h,
\end{equation}
where $\oplus$ denotes the concatenation and $\odot$ denotes the Hadamard product. 

\begin{figure*}[!h]
	\centering
	\includegraphics[width=.9\linewidth]{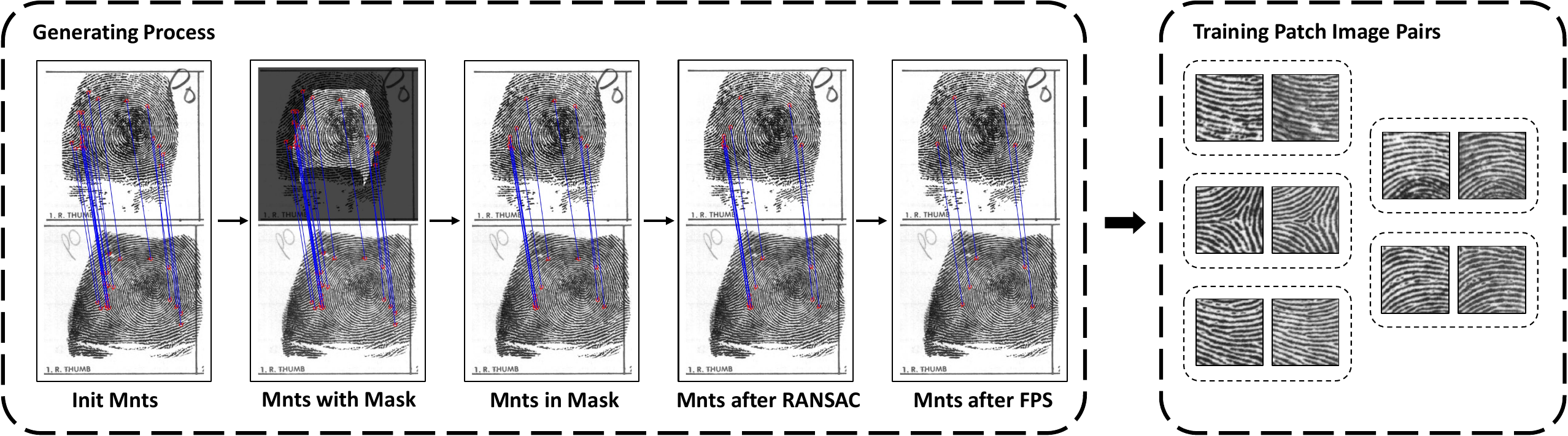}
	\caption{The process of selecting training minutiae pairs.}
	\label{fig:mnt_select}
\end{figure*}

\subsection{Training Loss} \label{sec:training_loss}
\textbf{Classification Loss}.
We incorporate the robust CosFace loss \cite{cosface} used in face recognition to refine the learning of fingerprint feature representations. This loss function is applied distinctly to both minutiae and texture descriptors. For each descriptor, the process begins with flattening, followed by passing through a fully connected (FC) layer that categorizes into $V$ classes. Here, $V$ also signifies the count of minutiae-centered training image pairs, a procedure detailed in Sec. \ref{sec:generate_sample}. 
The loss is calculated by 
\begin{equation}
   \label{eq:cosface}
   \mathcal{L}_\text{cls}^{i} = -\frac{1}{N} \sum_{n=1}^{N}\log \frac{e^{A(\cos(\theta_{y_n}^{i}) - b)}}{e^{A(\cos(\theta_{y_n}^{i}) - b)} + \sum_{v=1,v\neq y_n}^{V}e^{A\cos(\theta_{v}^{i})}},
\end{equation}
where $\cos(\theta_{v}^{i})$ is calculated by 
\begin{equation}
   \cos(\theta_{v}^{i}) = W_{v}^\mathrm{T}f_{i}, \quad W_v=\frac{W_v}{\|W_v\|},\quad f_i=\frac{f_i}{\|f_i\|}.
\end{equation}
 The $A$ is employed for normalizing magnitude, with $i$ signifying the type such that $i\in\{\text{t}, \text{m}\}$. The $W_v$ represents the learnable weight vector of the $v$-th class. The term $b$ represents the margin, $N$ quantifies the count of samples per batch, and $y_n$ indicates the class label of the sample $f_i$.

 \textbf{Segmentation Loss and Minutiae Loss}.
 We adopt the binary cross entropy loss for calculating segmentation loss $\mathcal{L}_{\text{seg}}$ and utilize mean square error for calculating minutiae loss $\mathcal{L}_{\text{mnt}}$. We utilize the VeriFinger v12.0 to extract the target minutiae and segmentation maps. 

 \textbf{Similarity Loss}.
 To ensure the local feature consistency within fingerprints from the same finger with different distortion or valid area, we simulate a counterpart plain fingerprint using the segmentation map from Diverse Pose Fingerprint dataset (DPF) \cite{Duan2023FingerPose} and a simulated fingerprint distortion field following the model \cite{Si2015DistortedFinger} according to the original rolled one. Similarity loss is to keep the corresponding region's feature of the rolled fingerprint similar to the feature of plain one. It is defined as 
 \begin{equation}
   \label{eq:sim}
   \mathcal{L}_\text{sim} = \frac{1}{|h_{p\cap r}|} \sum\nolimits_{(i,j)\in h_{p\cap r}} \| f_p^{ij} - f_r^{ij} \|^2,
 \end{equation}
 where $f_p$ and $f_r$ denote the representations extracted from plain and rolled fingerprints respectively. 
 
 Therefore, the overall supervision loss is defined as 
 \begin{equation}
   \label{eq:loss}
   \mathcal{L} = \sum_{i}^{\{\text{t},\text{m}\}} \mathcal{L}_{\text{cls}}^{i} + \lambda_{\text{seg}}\mathcal{L}_{\text{seg}} + \lambda_{\text{mnt}}\mathcal{L}_{\text{mnt}} + \lambda_{\text{mnt}}\mathcal{L}_{\text{sim}},
 \end{equation}
 where $\lambda_{\text{seg}}$, $\lambda_{\text{mnt}}$, and $\lambda_{\text{mnt}}$ are weight to balance the loss components.

\subsection {Training Sample Generation} \label{sec:generate_sample}
In contrast to the approach of directly selecting minutiae according to MCC \cite{cappelli2010minutia} minutiae pair matching scores as presented in \cite{MinNet}, our methodology incorporates a multitude of selection strategies. These strategies are designed to identify minutiae that not only are correctly matched but also resilient to distortion. Furthermore, they facilitate the selection of distinct regions for network training. This approach enables the network to learn distinctive features across varying fingerprint patches, enhancing its capability to differentiate between unique minutiae configurations. The training samples generation process is shown in Figure \ref{fig:mnt_select}.

\textbf{Extracting Minutiae}.
We utilize VeriFinger v12.0 \cite{nist2020verfinger} to extract minutiae from fingerprint images. Faced with a scarcity of available public latent fingerprints for our training needs, we resort to employing the once publicly available rolled fingerprint dataset NIST SD14 \cite{NIST14} as our training dataset.

\textbf{Mated Minutiae}. 
 We employ the Minutia Cylinder-Code (MCC) \cite{cappelli2010minutia} method to identify corresponding minutiae pairs in genuine fingerprint matches. However, it's important to note that not all identified minutiae pairs are accurate. Initially, we prioritize the top $K$ minutiae pairs based on their matching scores. Subsequently, considering the bad training quality of regions located at the edges of the fingerprint's foreground which is susceptible to erroneously identifying minutiae, We utilize the segmentation map obtained from the enhancement process conducted by VeriFinger v12.0, applying erosion, to exclude minutiae located in invalid regions. Moreover, we employ the RANSAC algorithm to compute a 2D affine transformation matrix that aligns the source minutiae with the target minutiae. This step facilitates the removal of incorrectly matched minutiae pairs as well as those affected by significant fingerprint distortions. 
Finally, we acknowledge that training images generated from closely situated minutiae often resemble each other, thus posing a challenge for differentiation during the training phase. To mitigate this issue, we implement Farthest Point Sampling (FPS) to judiciously choose a subset comprising a maximum of $K$ minutiae, which ensuring that the selected minutiae are spaced sufficiently far apart. We designated $N=12$ and $K=5$ for our training.

\textbf{Generate Image Samples}.
After the creation of matched minutiae pairs, we proceed to transform the training image samples. This transformation involves translating and rotating the fingerprint images to align with the position and orientation of each specific minutia, and then cropping. As a result, the transformed patch images are centered on the anchor minutiae, with these minutiae oriented horizontally to the right. In our application, we opted for a patch size of $128 \times 128$ pixels in 500 ppi.

\begin{table*}[!t] 
	\small
	\begin{center}
		\begin{threeparttable}
			\begin{tabular}{p{.12\linewidth}<{\centering}*{1}{p{.14\linewidth}<{\centering}}*{2}{p{.25\linewidth}<{\centering}}}
				\toprule
				\multirow{2}{*}{Dataset} 	
				& NIST SD14\tnote{$\dagger$}
				& THU Contact10K\tnote{$\ddagger$}
				&  DPF\tnote{$\dagger$} \\
				\cmidrule(lr){2-2} \cmidrule(lr){3-3} \cmidrule(lr){4-4}
				& Rolled
				& Rolled or Plain
				& Plain \\
				\midrule
				Image    & 
            \raisebox{-.5\height}{\includegraphics[height=.7in]{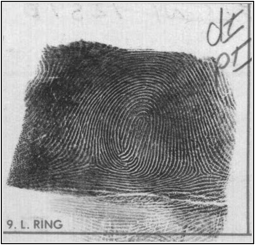}}
				&  \raisebox{-.5\height}{\includegraphics[height=.7in]{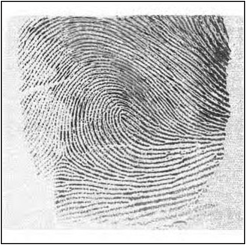}} ~\raisebox{-.5\height}{\includegraphics[height=.7in]{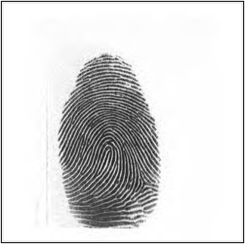}}
				& 
             \raisebox{-.5\height}{\includegraphics[height=.7in]{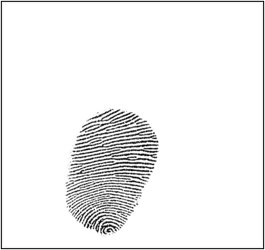}}~\raisebox{-.5\height}{\includegraphics[height=.7in]{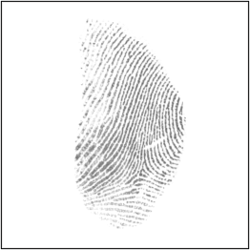}}\\
				\hhline
				Sensor
				& Inking                 
				& Inking / Optical
				& Optical \\
				\hhline
				Description              
				& 27,000 pairs
				& 10,458 fingerprints
				& 40,112 fingerprints \\
				\hhline
				Usage                     
				& Training
				& Testing
				& Augmentation \\
				\bottomrule
			\end{tabular}

         \begin{tabular}{p{.12\linewidth}<{\centering}*{3}{p{.18\linewidth}<{\centering}}*{1}{p{.20\linewidth}<{\centering}}}
				\toprule
				\multirow{2}{*}{Dataset} 	
				& \multicolumn{2}{c}{NIST SD27\tnote{$\dagger$}}
				& \multicolumn{2}{c}{N2N Latent} \\
				\cmidrule(lr){2-3} \cmidrule(lr){4-5}
				& Rolled
				& Latent
				& Rolled & Latent \\
				\midrule
				Image    & \raisebox{-.5\height}{\includegraphics[height=.7in]{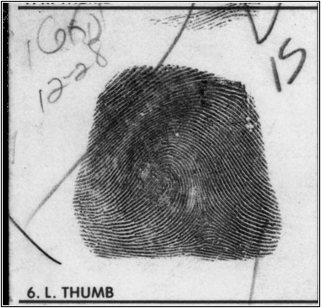}}
				& \raisebox{-.5\height}{\includegraphics[height=.7in]{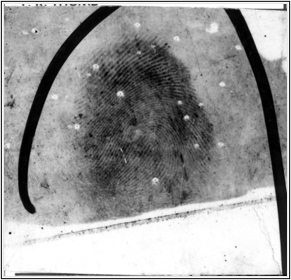}}
				& \raisebox{-.5\height}{\includegraphics[height=.7in]{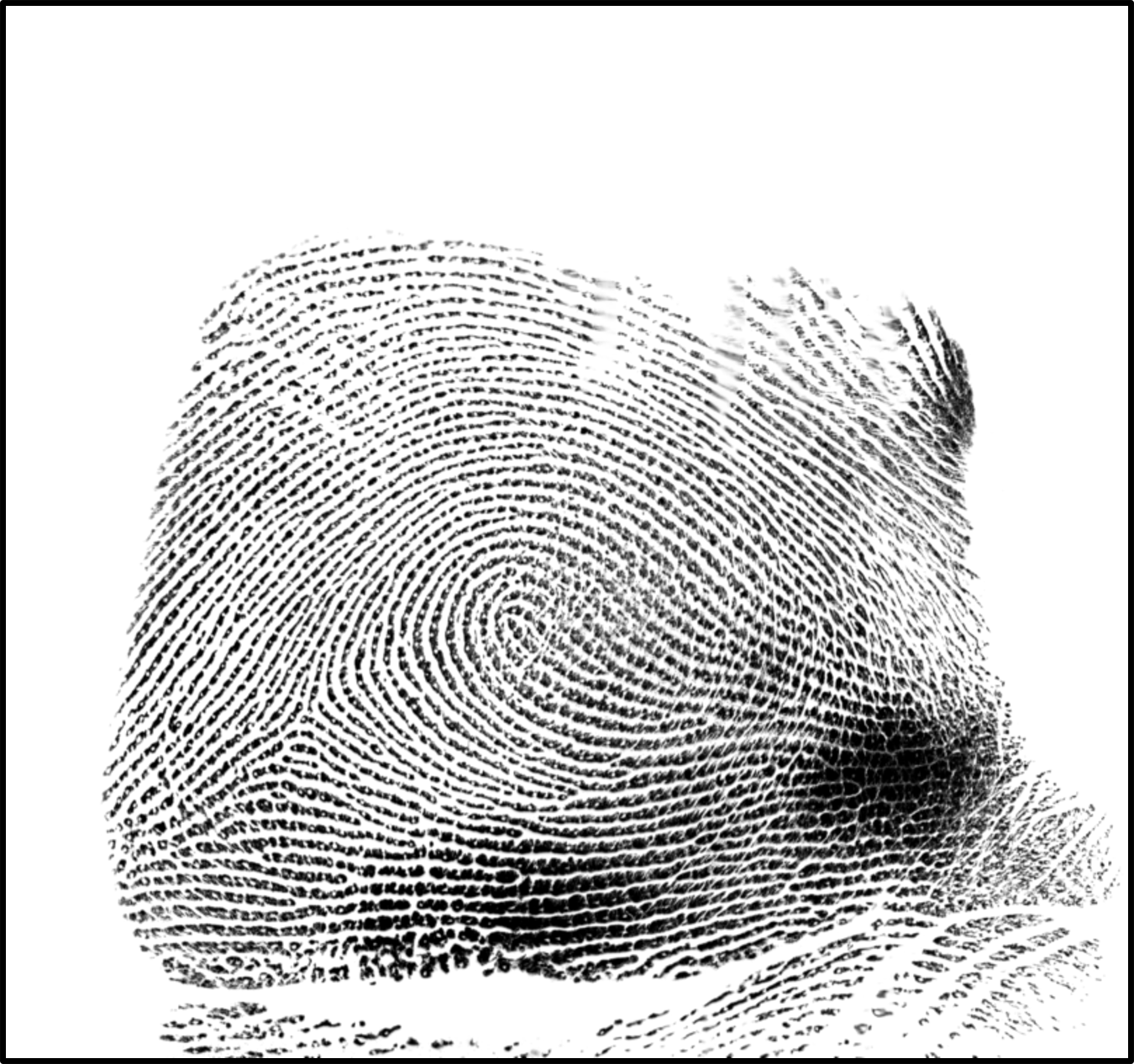}}
            &\raisebox{-.5\height}{\includegraphics[height=.7in]{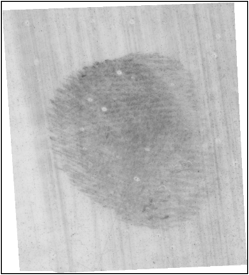}}
            \\
				\hhline
				Sensor
				& Inking  
            & ---             
				& Optical
				& --- \\
				\hhline
				Description              
				& \multicolumn{2}{c}{258 pairs}
				& 2,000 fingerprints
				& 3,318 fingerprints \\
				\hhline
				Usage                     
				& \multicolumn{2}{c}{Testing}
				& \multicolumn{2}{c}{Testing} \\
				\bottomrule
			\end{tabular}
         	\begin{tablenotes}
				\item[$\dagger$] Corresponding datasets are discontinued.
				\item[$\ddagger$] Corresponding datasets are internal.
			\end{tablenotes}
		\end{threeparttable}
	\end{center}
	\vspace{-0.1cm}
	\caption{All fingerprint datasets used in this study.}
	\label{tab:datasets}
\end{table*}

\subsection{Fingerprint Matching} \label{sec:fp_matching}
Our fingerprint matching process unfolds in two stages: calculating local similarities between two minutia sets from two fingerprint images; getting the final matching score of two images from the local similarity matrix. 

Initially, we compute the initial score matrix $S_{(A,B)}$ by comparing minutiae dense descriptors for the pair of fingerprints $(A, B)$ under comparison. The matching score for a pair of minutiae dense descriptors is determined through the cosine similarity between the two flattened descriptors. Subsequently, we adopt the score normalization technique outlined in \cite{Iris} to mitigate the impact of variations in the area of the overlapping region on the score. After getting the descriptors $({f_{a_i}}, {f_{b_j}})$ from two minutiae $(a_i, b_j)$ by Eq. \eqref{eq:generate_desc} and flattening them $({f_{a_i}}^\prime, {f_{b_j}}^\prime)$, the score is computed by
\begin{equation}
   \label{eq:score_calc}
   S_{(A, B)}(i,j) = \frac{\Braket{{f_{a_i}}^\prime, {f_{b_j}}^\prime}}{\|f_{a_i}\odot h_{b_j}\|_F~\|f_{b_j}\odot h_{a_i}\|_F} \cdot \sqrt{\frac{h_o}{H_o}},
\end{equation}
where $(a, b)$ represent the minutiae sets from fingerprints $(A, B)$, $h_o$ is the area of overlapping region $h_o = |h_{a_i\cap b_j}|$, and $H_o$ is a constant that reflecting the average of overlapping area. We set $H_o = 1326$ in our method. 

Subsequently, we apply the Local Similarity Assignment with Relaxation (LSA-R) method, as introduced in MCC \cite{cappelli2010minutia}, to derive the final matching score from similarity matrix $S_{(A,B)}$ and minutiae sets $(a, b)$. The LSA-R method addresses the linear assignment problem using $S_{(A,B)}$ through a combination of the Hungarian algorithm and a relaxation approach that takes into account the geometric configuration of the minutiae sets $(a, b)$ \cite{Feng2006FingerMatch}. Then we get the adjusted score matrix $S^\prime_{(A,B)}$, and get the top $n_m$ matching scores related to the minutiae sets $m={\{(a_i, b_i), i=1,...,n_m\}}$.
The $n_m$ is calculated as
\begin{equation}
	\label{eq:n_Set}
	n_m = min_{n_m} + \lfloor \frac{max_{n_m}-min_{n_m}}{1+e^{(-\tau(\text{min}(n_a,n_b)-\mu))}} \rceil,
\end{equation}
$(n_a, n_b)$ represents the number of minutiae sets $(a,b)$. We set the hyper parameters as $min_{n_m}=4, max_{n_m}=12, \tau=0.4,\mu=20$. $\lfloor\cdot\rceil$ is the rounding operator.
Finally, the matching score $\Gamma(A,B)$ between fingerprints $(A, B)$ is calculated as 
\begin{equation}\label{eq:final_score}
	\Gamma(A,B) = \frac{\sum_{(r,c)\in m}S^\prime_{(A,B)}(r,c)}{n_m}.
\end{equation}

\noindent\textbf{Implementation Details}. 
 To enhance data diversity, we generate plain fingerprints by cropping rolled fingerprints using segmentation maps from DPF dataset as described in Similarity Loss of Sec. \ref{sec:training_loss}. Minutiae outside the segmentation map's effective area are excluded. Augmentation also includes random translations up to 10 pixels, rotations within the range of $[-5^\circ, 5^\circ]$, intensity transformation, distortions, and Gaussian noise. We apply the distortion model by Si et al. \cite{Si2015DistortedFinger} to generate distortion fields. For optimal performance and computational efficiency, we configure the descriptors dimension $C$ to 6 and adjust the parameters $A$ and $b$ in Eq. \eqref{eq:cosface} to 30 and 0.4, respectively. The parameters $\lambda_{\text{seg}}$, $\lambda_{\text{mnt}}$, and $\lambda_{\text{mnt}}$ in Eq. \eqref{eq:loss} are set to 1, 0.01, and 0.00125, respectively. We employ the AdamW optimizer with a learning rate of $3.5\times 10^{-4}$, and apply L2 regularization to the trainable parameters to prevent overfitting. 

\begin{figure*}[!ht]
	\centering
	\subfloat[CMC curves of NIST SD27\label{fig:cmc_nist27}]{\includegraphics[height=.3\linewidth]{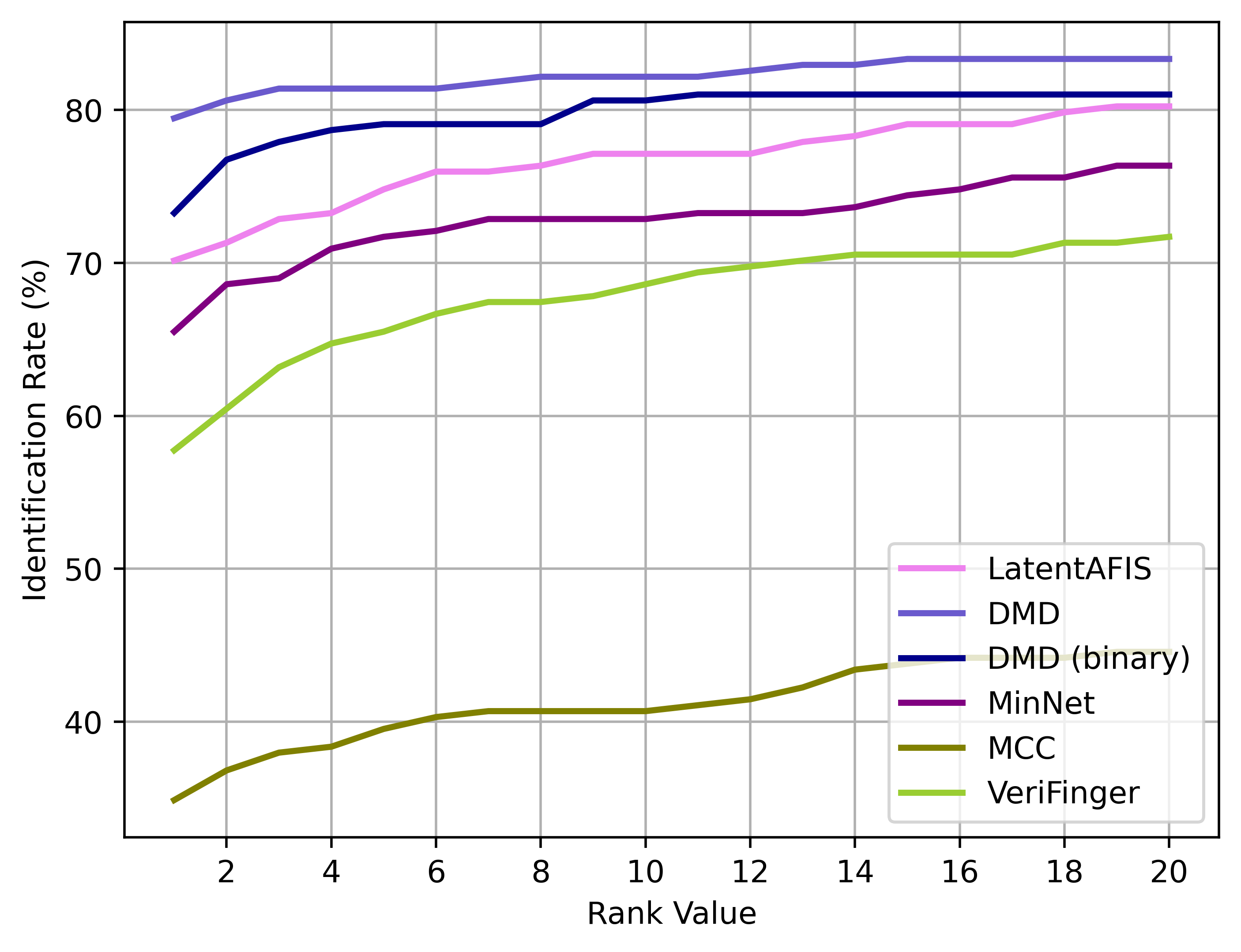}} \hfil
	\subfloat[CMC curves of N2N Latent\label{fig:cmc_n2nlatent}]{\includegraphics[height=.3\linewidth]{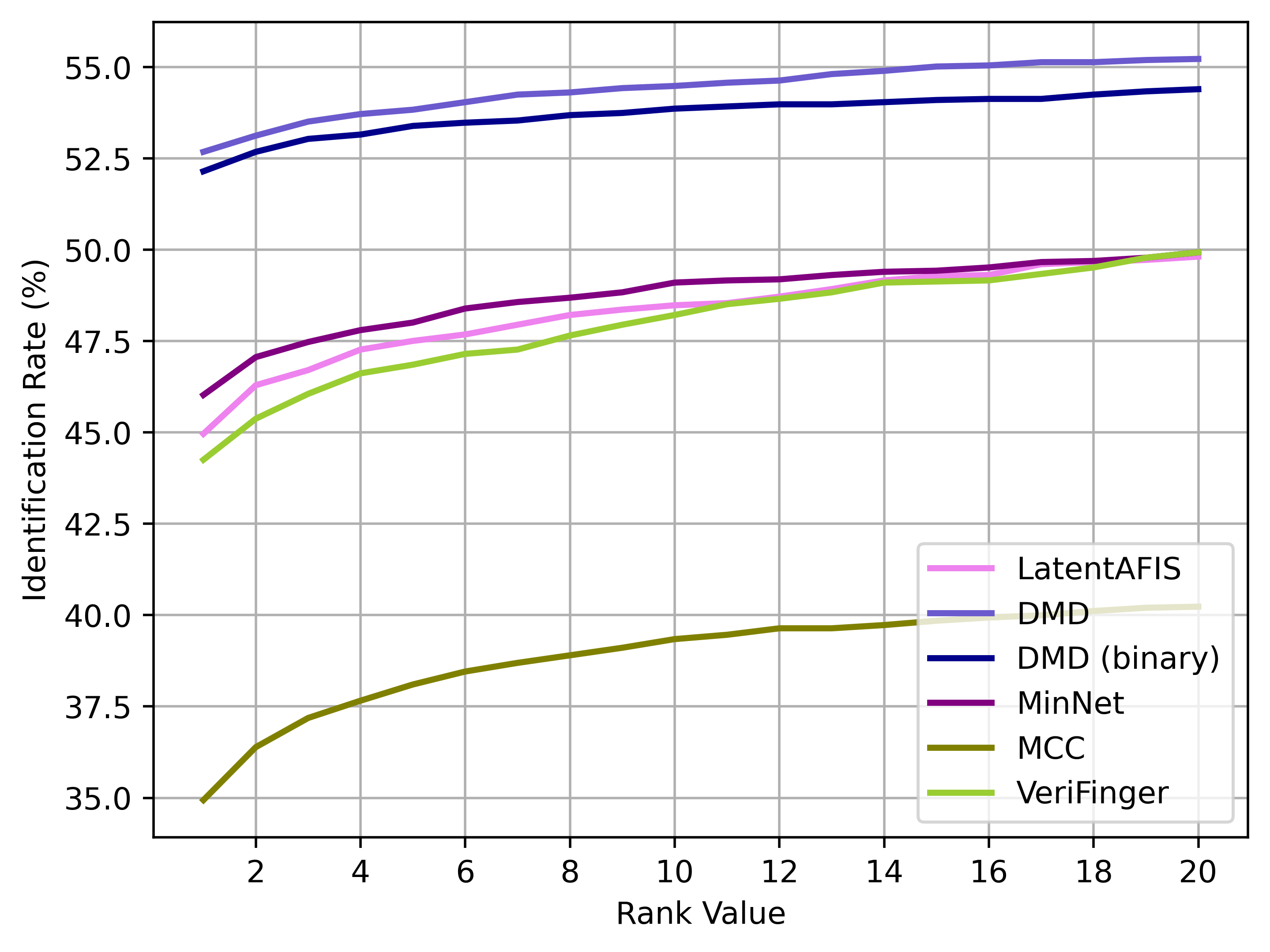}} \hfil
	\subfloat[DET curves of NIST SD27\label{fig:det_nist27}]{\includegraphics[height=.3\linewidth]{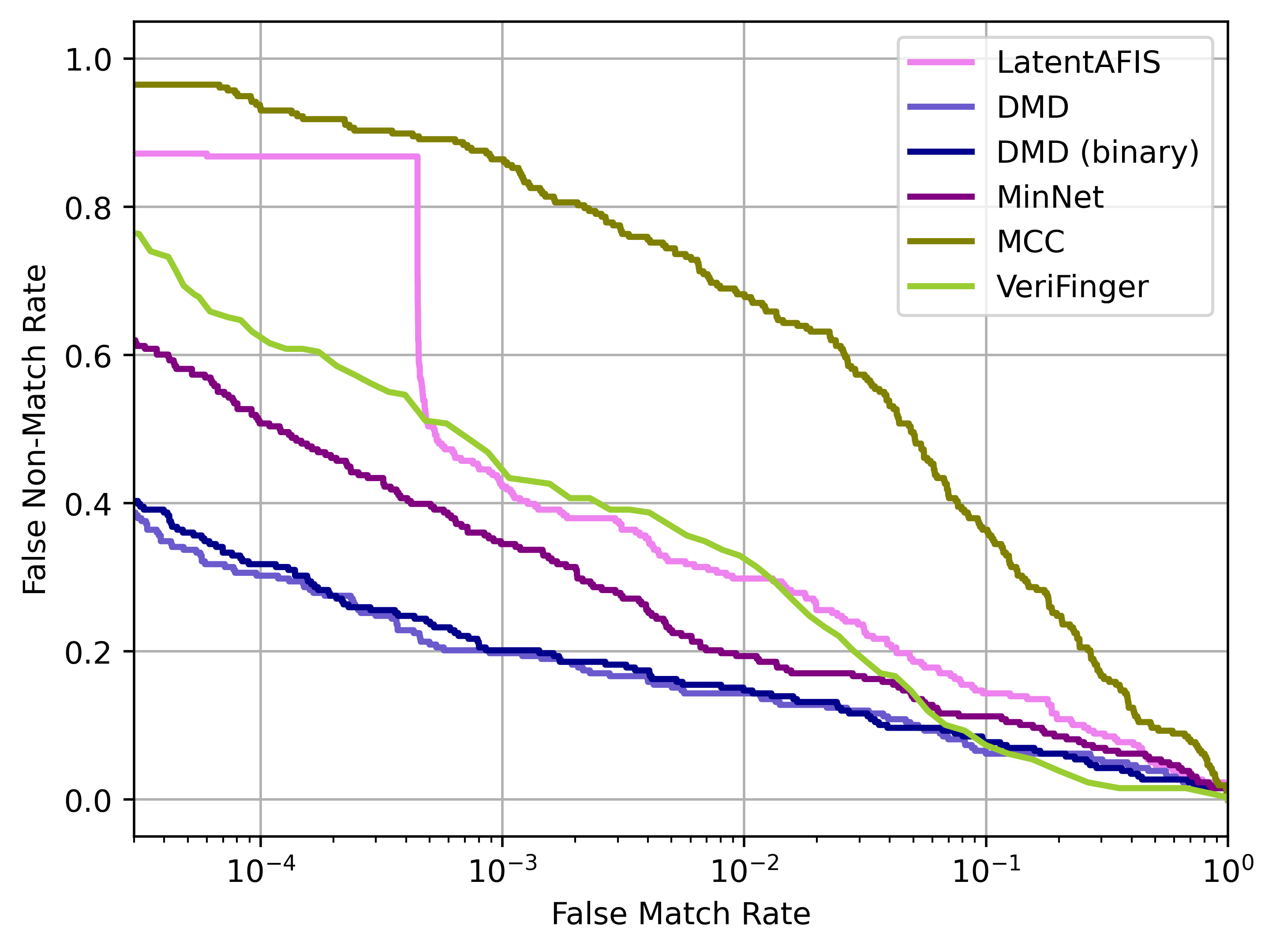}} \hfil
	\subfloat[DET curves of N2N Latent\label{fig:det_n2nlatent}]{\includegraphics[height=.3\linewidth]{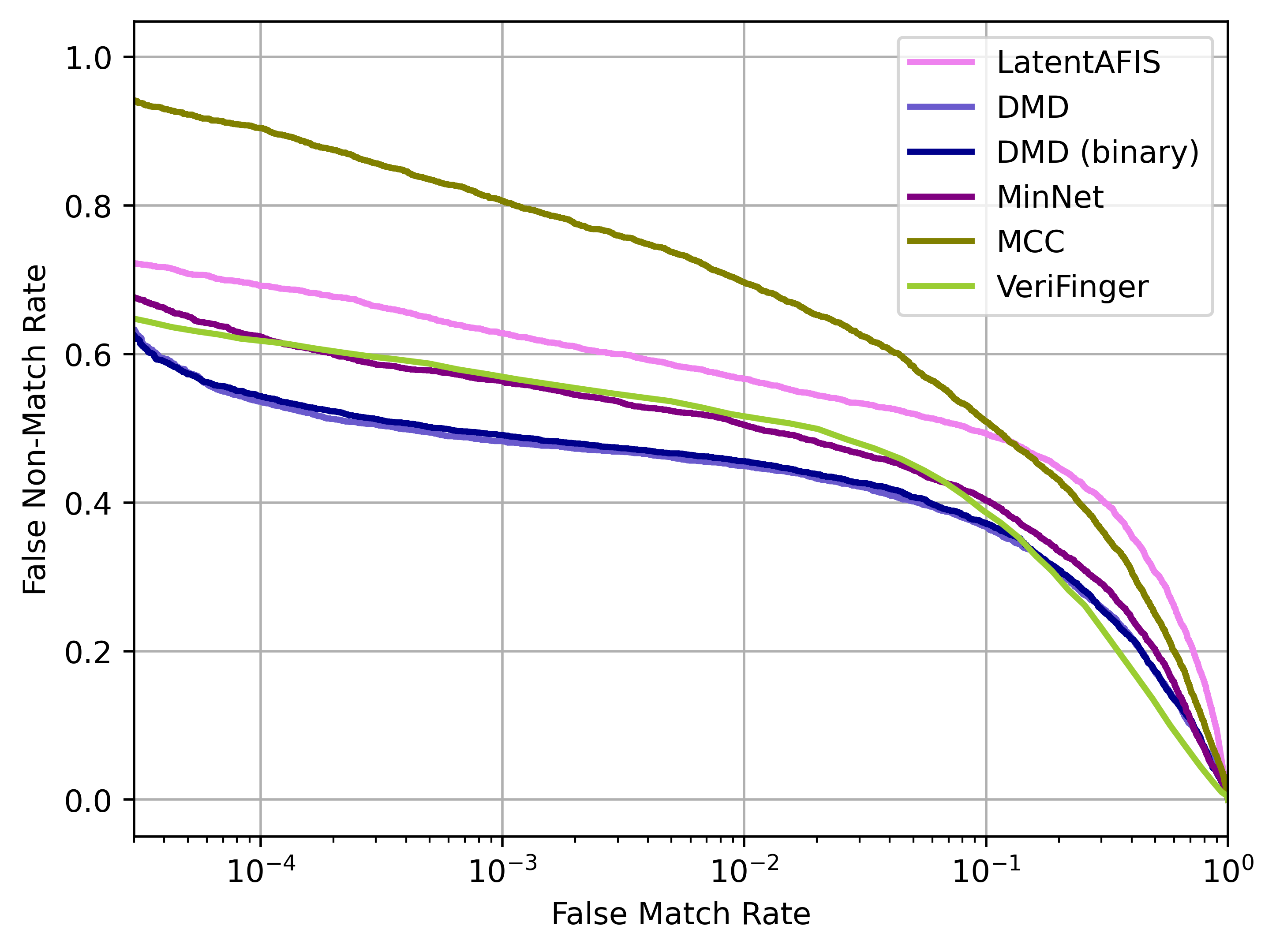}}
	\vspace{0.05cm}
	\caption{Latent fingerprint matching performance on NIST SD27 (a) (c) and N2N Latent (b) (d).}
	\label{fig:quantative_results}
\end{figure*}

\section{Experiment}
\subsection{Datasets}
In our study, we primarily rely on five datasets for both training and evaluation. Table \ref{tab:datasets} presents an overview of these datasets, including examples of fingerprints from each. For the training phase, we employ the NIST SD14 rolled fingerprint dataset, generating a total of 132,550 pairs of minutiae-centered patch fingerprints from its 27,000 pairs of fingerprints. Diverse Pose Fingerprint dataset (DPF) \cite{Duan2023FingerPose}, encompassing 776 fingers captured in a variety of poses, is utilized to simulate plain images for data augmentation. It is important to note that we do not engage in any fine-tuning on latent fingerprints for evaluation purposes. The NIST SD302 (N2N) dataset comprises fingerprints from 200 individuals, totaling 2,000 fingers. In particular, subset U serves as our gallery. Following the selection criteria detailed in \cite{gu2022latent}, we choose 3,383 latent fingerprints with reasonable quality out of 10,000 available. Additionally, the NIST SD27 dataset includes 258 latent-to-rolled fingerprint pairs. To expand the gallery of NIST SD27, we incorporate a private dataset denoted as THU Contact10K dataset, which contains 10,458 rolled or plain fingerprints from ten fingers of over 1046 different subjects. 

\begin{figure*}[!h]
	\centering 
	\includegraphics[width=.80\linewidth]{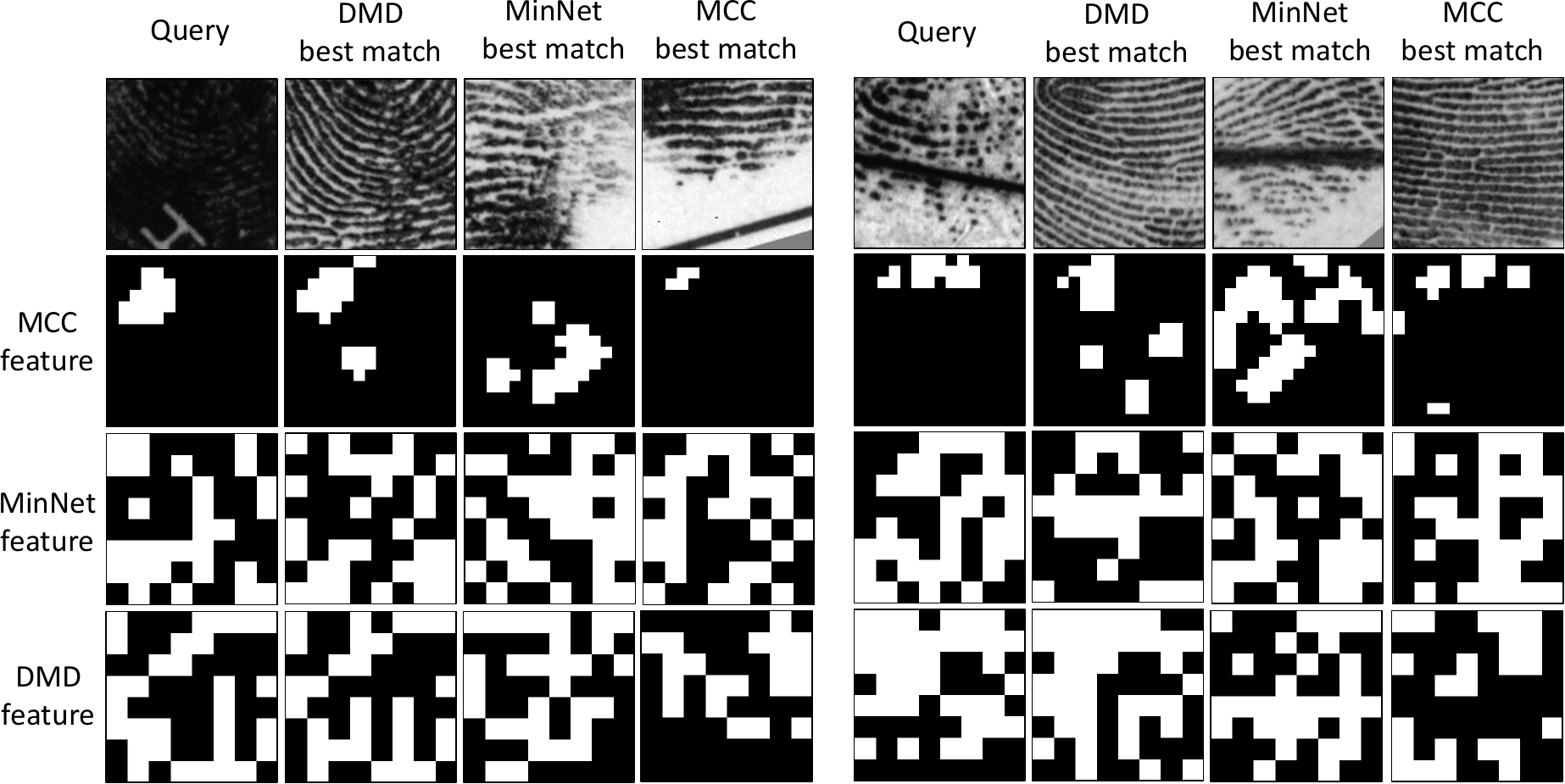}
	\vspace{0.05cm}
	\caption{Descriptor visualization on different patch images. Descriptors of MinNet are resized to three-dimension form for visualization. We select a specific channel from the aforementioned descriptors and convert it to a binary format to enhance visualization.}
	\label{fig:feature_visualization}
\end{figure*}

\begin{table}[!t]
	\small
	\begin{center}
	\begin{tabular}{*{3}{l}}
		\toprule
		\textbf{Type} & \textbf{Approach} & \textbf{Setting} \\
		\midrule
		\multirow{2}{*}{Deep Learning} & MinNet \cite{MinNet} & Our reimplementation \\
		\cmidrule(lr){2-3} & LatentAFIS \cite{Cao2020E2E} & Original public code \\
		\midrule
		Convention & MCC \cite{cappelli2010minutia} & Our reimplementation \\
		\hhline
		COTS & VeriFinger \cite{nist2020verfinger} & Commercial SDK \\
		\bottomrule
	\end{tabular}
	\end{center}
	\vspace{-0.1cm}
	\caption{Experiment settings of approaches to be compared.}
	\label{tab:compared_methods}
\end{table}

\subsection{Compared Methods}
To ascertain the effectiveness of Dense Minutia Descriptor (DMD), we conduct a comprehensive comparison with both traditional and more recent deep learning minutiae-based fingerprint recognition methods (Table \ref{tab:compared_methods}). Specifically, our evaluation of LatentAFIS \cite{Cao2020E2E} utilize the publicly available code and the released model weights provided by the authors. We retain the entire pipeline of their system, which includes image enhancement, the extraction of minutiae and virtual minutiae templates, as well as the processing of descriptors. Regarding MinNet \cite{MinNet}, we train  it with original patch fingerprint images instead of enhancement ones which are the same as ours, and increase the dimensionality of its descriptors to 768 to align with our configuration. As to commercial matcher VeriFinger v12.0, it has two types of matchers: ISO minutia-only template and proprietary template consisting of minutiae and other features. To make a high baseline, we adopt the proprietary template which presents higher performance. And we reimplement MCC \cite{cappelli2010minutia} to achieve a faster extraction and matching speed while maintaining the same matching performance as its public SDK. The minutiae of testing fingerprint images were extracted using VeriFinger and thus identical across these methods, with the exception of LatentAFIS, which extracts its minutiae and templates using the model weights provided in its release. Fingerprint matching process for MCC, MinNet, and DMD follows the same procedure, as detailed in Sec. \ref{sec:fp_matching}, allowing a fair comparison of different minutia descriptors. And the matching strategies for VeriFinger and LatentAFIS adhere to the protocols established by their respective systems.

\begin{figure}[!t]
	\centering 
	\includegraphics[width=\linewidth]{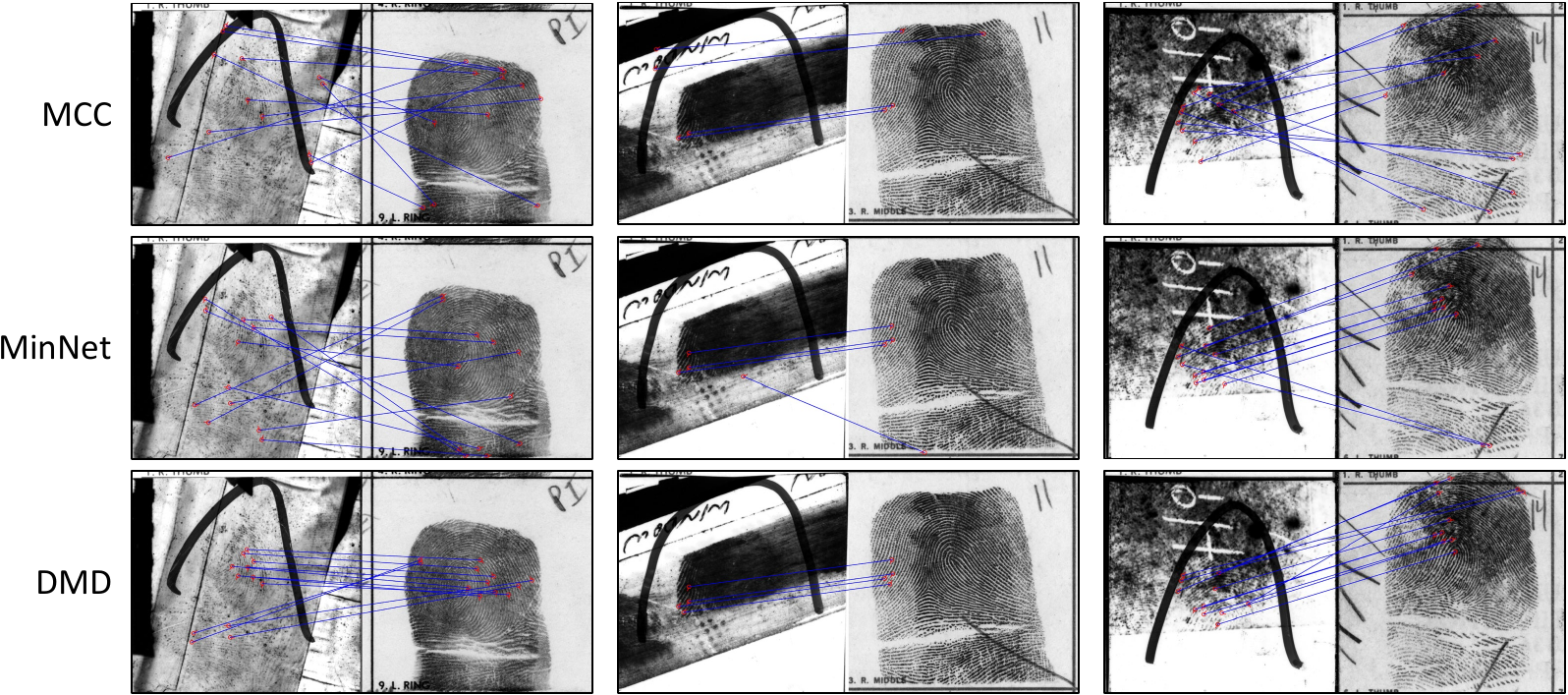}
	\caption{Top $n_m$ minutiae patch matching of three genuine pairs via different methods. $n_m$ is determined by Eq. \eqref{eq:n_Set}.}
	\label{fig:matching_visualization}
\end{figure}

\begin{table}[!t]
	\small
	\begin{center}
		\begin{threeparttable}
			\begin{tabular}{{p{.25\linewidth}}*{4}{p{.13\linewidth}<{\centering}}}
				\toprule
				\multicolumn{1}{c}{\multirow{2}{*}{\textbf{Method}}}  &
				\multicolumn{2}{c}{\textbf{NIST SD27}} &
				\multicolumn{2}{c}{\textbf{N2N Latent}} \\
				\cmidrule(lr){2-3} \cmidrule(lr){4-5} &
				\textbf{Rank-1} & \textbf{TAR} & \textbf{Rank-1} & \textbf{TAR}\\
				\midrule
				MCC \cite{cappelli2010minutia} & 35.27 & 13.57 & 34.94 & 19.42 \\
				VeriFinger \cite{nist2020verfinger} & 53.10 & 58.91 & 44.31 & 42.71 \\
				LatentAFIS \cite{Cao2020E2E} & 70.16 & 57.75 & 44.96 & 37.22 \\
				MinNet \cite{MinNet} & 65.89 & 65.50 & 46.02 & 43.63 \\
				\hhline
				DMD (binary) & \underline{73.26} & \underline{79.84} & \underline{52.14} & \underline{50.90} \\
				DMD & \textbf{79.07} & \textbf{80.23} & \textbf{52.68} & \textbf{51.73} \\
				\bottomrule	
			\end{tabular}
		\end{threeparttable}
	\end{center}
	\vspace{-0.1cm}
	\caption{Verification and recognition accuracy (\%) on latent fingerprint datasets. TAR@FAR=0.1\% is reported.}
	\label{tab:quantitive_results}
\end{table}

\subsection{Latent Fingerprints Matching Performance}
The proposed DMD is benchmarked against the array of methodologies as delineated in Table \ref{tab:compared_methods}. Rank-1 and TAR@FAR=0.1\% metrics, alongside the Cumulative Match Characteristic (CMC) curve and the Detection-Error Tradeoff (DET) curve, serve as crucial tools for the quantitative evaluation of various methods. Moreover, we extend our work to include a binary variant of DMD, wherein $\bar{f}_{\text{t},\text{m}}$ is binarized to 0 and $h$ is thresholded at 0.5 for binarization. It culminates in a more streamlined version of the DMD, with each descriptor being condensed to occupy merely 96 bytes.

Compared to other minutia-based fingerprint matching methods, DMD stands out by a significant margin in terms of matching and indexing capabilities, as demonstrated in Table \ref{tab:quantitive_results} and Figure \ref{fig:quantative_results}. This superiority continues to hold even when compared to binary DMD. 
The exceptional performance of DMD can be attributed to its remarkable representation. Figure \ref{fig:feature_visualization} showcases three primary types of descriptors: spatial representation derived from minutia distribution (MCC), one-dimensional representation consisting of abstract features (MinNet), and spatial representation modeled from texture and minutiae information using abstract features (DMD).

One-dimensional descriptors, which do not closely correlate with the spatial characteristics of original fingerprints, may face challenges. These include difficulty in isolating the impact of noise present in latent fingerprints, which can affect the entire descriptor, as well as limitations in the interpretability of the descriptor itself. And we can observe that it exhibits an irregular pattern across features of these samples in Figure \ref{fig:feature_visualization}. MCC maintains a spatial relationship with the fingerprint's plane, yet it primarily models the distribution of minutiae in the vicinity, making it highly susceptible to inaccuracies caused by erroneously detected or missing minutiae. DMD not only retains the spacial representation like MCC but also uses the robust abstract features. From Figure \ref{fig:feature_visualization}, it can be observed that the DMD feature pattern of the query in the first example closely resembles its best match, except for the lower left part which is affected by the symbol ``H". Similarly, in the second example, the upper region of the matching pair set exhibits a strong resemblance in DMD features, despite the incomplete bottom region of the searched latent fingerprint and interference from a black line.

Moreover, the top $n_m$ patch matching selected by feature similarity of DMD is more accurate than others which is illustrated in Figure \ref{fig:matching_visualization}. This indicates that the value of the matching score can, to a great extent, determine whether the match is genuine. This also facilitates a better understanding of our strong performance on the TAR metric and the DET curve. It indicates the potential of DMD used in large-scale fingerprint indexing.

\subsection{Ablation Study}
In this section, we explore the impact of various modifications made to our proposed DMD, which include omitting the normalization approach outlined in Eq. \eqref{eq:score_calc}, decreasing the DMD dimensionality from $C=6$ to $C=3$, and merging two branches into a single one with segmentation head and minutiae head. It is noteworthy that the descriptor derived from a singular branch retains the dimensionality equivalent to the dual-branch with $C=3$, implying that $f\in\mathbb{R}^{6\times8\times8}$. The quantitative outcomes of these modifications are summarized in Table \ref{tab:ablation_results}.

We can observe that a normalization strategy, which takes the overlapping region into account, significantly improves the DMD matching performance. It effectively addresses the common scenario of low genuine match overlap areas in latent fingerprint matching, hence significantly improving the Rank-1 metric. Besides, DMD's performance deteriorates as dimensionality reduces ($C=3$), yet it still outperforms the single-branch one of the same dimensionality. Therefore, the dual-branch design, integrating different features (texture feature and minutiae feature), greatly enhances the model's performance.

\begin{table}[!t]
	\small
	\begin{center}
		\begin{tabular}{{p{.25\linewidth}}*{4}{p{.13\linewidth}<{\centering}}}
			\toprule
			\multicolumn{1}{c}{\multirow{2}{*}{\textbf{Modification}}}  &
			\multicolumn{2}{c}{\textbf{NIST SD27}} &
			\multicolumn{2}{c}{\textbf{N2N Latent}} \\
			\cmidrule(lr){2-3} \cmidrule(lr){4-5} &
			\textbf{Rank-1} & \textbf{TAR} & \textbf{Rank-1} & \textbf{TAR}\\
			\midrule
			w/o Norm & 76.74 & 79.07 & 49.28 & 49.13 \\
			$C=3$ & 75.58 & 78.68 & 48.57 & 48.30 \\
			Single Branch & 66.28 & 68.99 & 48.45 & 48.06 \\
			\hhline
			None & \textbf{79.07} & \textbf{80.23} & \textbf{52.68} & \textbf{51.73}\\
			\bottomrule	
		\end{tabular}
	\end{center}
	\vspace{-0.1cm}
	\caption{Ablation study on the verification and recognition accuracy (\%) of DMD. TAR@FAR=0.1\% is reported.}
	\label{tab:ablation_results}
\end{table}

\section{Limitation and Future Works}
Despite good performance of our proposed Dense Minutia Descriptor (DMD) on serveral latent fingerprint datasets, there remains room for enhancement in several aspects. The effectiveness of DMD heavily relies on the accuracy of the preceding minutiae extraction processes. In this study, we utilize VeriFinger v12.0 \cite{nist2020verfinger} for minutiae extraction. Although VeriFinger excels at identifying minutiae in medium or high quality fingerprints, its performance is compromised by many latent fingerprints with complex background patterns, often resulting in the extraction of incorrect minutiae from the background areas (Figure \ref{fig:matching_visualization}). Furthermore, in datasets such as NIST SD27 or N2N Latent, VeriFinger sometimes fails to extract a sufficient number of minutiae for effective matching. Thus, the potential for improving DMD may lie in leveraging a more robust latent fingerprint minutiae extractor or refining the selection of minutiae from existing tools with a precise foreground mask.

Secondly, our current approach does not incorporate the use of enhanced fingerprints as input. Insights from Grosz et al. \cite{grosz2023latent} suggest that the efficacy of latent fingerprint matching methods can significantly benefit from a tailored latent fingerprint enhancement technique. Motivated by this understanding, we aim to develop a bespoke latent fingerprint enhancement method that is specifically designed to improve the performance of DMD.

Given scenarios where the minutiae extractor fails to retrieve an adequate number of minutiae for effective matching, employing virtual minutiae (derived from the orientation field) \cite{Cao2017AutoLatent,Cao2018TextureTemplate,Cao2020E2E} as anchor points can be a viable solution. By adopting such approach, we have the potential to further enhance the matching performance of DMD in latent fingerprints. Moreover, this methodology could also prove advantageous for the matching of small fingerprints collected from smartphones or other mobile devices. 

\section{Conclusion}
	In this study, we introduce a deep network based dense minutia descriptor named DMD. This descriptor is presented as a three-dimensional construct, where two dimensions are aligned with the original image plane, and the third dimension encapsulates robust abstract features. To refine and enhance the representational capacity of DMD, we employ a strategic selection of training samples alongside a dual-branch architecture for its training. Additionally, the feature visualization sheds light on its interpretability within the context of fingerprint matching. We conducted evaluations of DMD against other contemporary methods using the NIST SD27 and N2N Latent datasets. The results demonstrate that DMD significantly outperforms competing methodologies in terms of Rank-1 identification rate and True Acceptance Rate (TAR) metrics. Remarkably, DMD maintains good matching performance even after undergoing a straightforward binarization process, which contributes to improved matching efficiency and the potential for secure template encryption. 

{\small
\bibliographystyle{ieee}
\bibliography{DMD}
}

\end{document}